# A Data-Driven Multi-Objective Approach for Predicting Mechanical Performance, Flowability, and Porosity in Ultra-High-Performance Concrete (UHPC)

**Chakma Jagaran [a], ZhiGuang Zhou [a,\*], Chakma Jyoti [b], Cao YuSen [a]**

[a] State Key Laboratory of Disaster Mitigation for Structures, Tongji University, Shanghai 200092, China
[b] Department of Electrical and Electronic Engineering, University of Chittagong, Chittagong-4331, Bangladesh

**Abstract**

This study presents a data-driven, multi-objective approach to predict the mechanical performance, flow ability, and porosity of Ultra-High-Performance Concrete (UHPC). Out of 21 machine learning algorithms tested, five high-performing models are selected, with XGBoost showing the best accuracy after hyperparameter tuning using Random Search and K-Fold Cross-Validation. The framework follows a two-stage process: the initial XGBoost model is built using raw data, and once selected as the final model, the dataset is cleaned by (1) removing multicollinear features, (2) identifying outliers with Isolation Forest, and (3) selecting important features using SHAP analysis. The refined dataset as model 2 is then used to retrain XGBoost, which achieves high prediction accuracy across all outputs. A graphical user interface (GUI) is also developed to support material designers. Overall, the proposed framework significantly improves the prediction accuracy and minimizes the need for extensive experimental testing in UHPC mix design.

∗Corresponding author: ZhiGuang Zhou,
E-mail addresses:
chakmajagaran@tongji.edu.cn (J. Chakma),
zgzhou@tongji.edu.cn (Z. Zhou),
jyotichakma101@gmail.com (J. Chakma),
2210050@tongji.edu.cn (Y. Cao)

## 1. Introduction

Ultra-High-Performance Concrete (UHPC) is an advanced composite material that is superior to normal concrete in terms of workability, durability, and mechanical properties [1]. With a compressive strength of higher than 150 MPa, a flexural strength over 50 MPa, and a tensile strength of more than 8 MPa, which is over 16 times higher than regular concrete [2]. This exceptional performance is the result of UHPC's meticulously designed composition, which combines evenly dispersed micro-steel fibers with a dense and refined microstructure with a low water-to-binder ratio (usually less than 0.25). The material's homogeneity, mechanical strength and durability are all improved by the optimal particle packing, which also reduces porosity [3]. Additionally, even under extreme climatic circumstances, the addition of high-strength fibers like steel fibers, which greatly improves toughness and resilience to external pressures by offering greater crack-bridging capabilities [4]. Critical infrastructure development, maintenance, and rehabilitation are just a few of the engineering applications that have made extensive use of UHPC owing to its remarkable strength and endurance [5]. It is very useful for both new construction projects and for strengthening the structure of existing buildings because of its high bonding capacity, pseudo-strain hardening behavior, and ability to sustain structural integrity after cracking [6]. Plus, the solid matrix of UHPC significantly reduces permeability, improving long-term resilience to environmental deterioration and chemical assaults [7].

Nowadays, UHPC mixes are being developed due to rigorous experimental study on mixture design factors. A study on lightweight sand discovered that it maximizes compressive and flexural strength while minimizing porosity [8]. The best river sand-to-lightweight sand (LWS) replacement ratio was 25%, with top compressive strength of 168 MPa after 91 days and top flexural strength of 24 MPa after 28 days [9]. Additionally, substituting quartz sand with glass sand decreased compressive strength by 10% [10]. Two further studies found that 18% Ceram site sand and steel fibers increased UHPC's interfacial transition zone, pore structure, mechanical strength, and specific strength [11]. Flexural strengths study shows steel fibers increase load-carrying capacity, energy absorption, and fracture resistance [12]. Likewise, it has been shown that adding a range of steel fibers and synthetic fibers increases UHPC's flexural and tensile strength [13]. Experimental testing was time- and labor-intensive, but it established dependability between the mixture design variable and material properties, according to past studies [14].

However, typical experimental methodologies are complicated, particularly with so many components. These elements include water, cement, fly ash, silica fume, nano-silica, slag powder, and limestone powder [15]. The type and content of aggregates, chemical admixtures such superplasticizers, and fibers like steel fibers and polymer fibers complicate mix design [16]. Cement-based components are the most variable of these materials and need long drying periods (usually more than 28 days) [17]. This usually requires many long trial-and-error experimentations. In addition, different materials must be employed to meet UHPC performance criteria in real-world engineering applications. As a result, it is time-consuming and inefficient to build UHPC mixes only using conventional experimental techniques.

In recent years, opposed to depending on conventional experimental testing, various research has adopted data-driven approaches, which have shown to be robust and useful tools for predicting the mechanical characteristics of UHPC. For instance, a four-layer artificial neural network trained with 927 data points predicted UHPC compressive strength using 18 mixed design parameters [18]. An artificial neural network with one hidden layer was trained with 162 data points to predict compressive strength utilizing mixture design elements like water-to-binder ratio and fiber content [19]. Additionally, a quadratic regression model with 12 data points predicted UHPC flexural strength [20]. Many studies have predicted mini-slump and porosity. Based on experimental data from six materials and 43 test sets, another study estimated compressive strength and mini-slump using ANN and statistical mixture design [21]. Recent studies have used hyperparameter-based machine learning methods to predict UHPC properties. A notable study used target optimization to build stronger, more fluid UHPC [22]. Moreover, optimization methods have been widely applied to construct UHPC with multi-objective performance feature [23]. Multi-objective optimization methods have also been effectively used in a variety of domains, exhibiting their resilience and applicability [24], [25].

Although this good performance, data-driven models still have significant drawbacks. (1) Overfitting in the small dataset makes data collection difficult, reducing generalization and prediction accuracy. (2) Current models have limited applicability. Insufficient input variable dimensionality reduces UHPC material efficiency and prediction accuracy. (3) High data variability increases learning costs, making parameter optimization difficult and requiring more complex training models. Machine learning algorithm training noise from structured and unstructured outliers lowers prediction accuracy. Finding and deleting outliers is hard. (4) Hyperparameter tuning improves model performance. Random search optimization method is using which enhance faster generalization, compare to others. (5) Most studies have predicted single performance outcomes based on component materials, with little research on multi-objective prediction on optimizing designs for varied UHPC mix proportions. (7) Few studies have estimated UHPC flowability, flexural, tensile, and porosity. The above barriers limit data-driven UHPC development and applications.

The remainder of this paper is structured as follows in order to achieve the aforementioned goals: Section 2 provides a detailed machine learning methodology and overview of the framework. Section 3 interprets the creation and enhancement of the dataset. Section 4 discusses the effectiveness of machine learning models in predicting key characteristics of UHPC, as well as the contribution of mixture design factors in the multi-objective optimization-based UHPC model. Section 5, the findings of the research outline its limitations and future scope.

## 2. Methodology

The methodological framework is presented in this section. The provided machine learning framework, which integrates many cutting-edge approaches and is clarified in Sections 2.2 to 2.6, is outlined in Section 2.1.

## 2.1 Overview of the work

The study develops a numerous machine learning framework to predict UHPC performance based on mixture ratio design and material attributes according to the target performance criteria. Fig. 1 demonstrates the flowchart of the frameworks with the eight primary steps: (1) Development and preparation of datasets: Utilize computer-generated methods for collecting experimental structure and unstructured data from an extensive amount of literature. Organized data, like tabular data, is referred to as structured data. Graphs or other unstructured data are unregulated. (2) The collected data are standardized and divided into training and test datasets. Section 2.2, interpret a plot digitizer approach to extract unstructured data and the process for improving and preparing datasets. (3) A variety of machine learning techniques are extensively evaluated on a range of datasets, and the most promising models are optimized to improve their prediction capabilities. An in-depth review of standard performance indicators often used in machine learning model assessment tells the final selection. Section 2.3 presents a predictive model selection which introduces a selection of 2.3.1 machine learning models with 2.3.2 the performance indicators. (4) To find the outlier data, the contamination ratio has been used. The selection of most contributed mixtures after SHAP value analysis. Section 2.4 and 2.5 explains the outlier detection approach and SHAP interpretation analysis. (6) Hyperparameter tuning is performed for the selected machine learning method based on K-fold cross-validation and random search. Section 2.6 presents the K-fold cross-validation method and the random search of hyperparameter tuning. The evaluation metrics such as mean absolute error (MAE), percentage mean absolute error (PMAE), mean squared error (MSE), root mean square error (RMSE), maximum absolute error (MaxAE), and coefficient of determination (R2) are used to evaluate the performance of the model. Interpretations of these metrics reveal how mixture design features influences on UHPC properties.

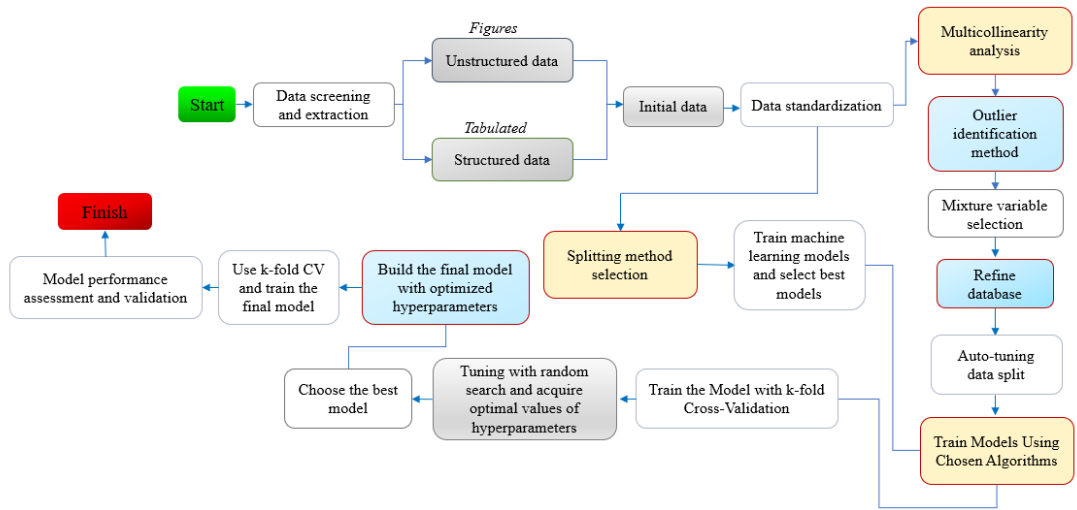

**Fig.1**
presents the workflow of the data-driven model designed to predict UHPC characteristics.

## 2.2 Development and preparation of datasets

In order to ensure the precision and effectiveness of data-driven research, dataset development and preparation are crucial processes. The unstructured data is less often used in data-driven research nowadays because of its usually long extraction times. Therefore, this research suggests a four-step semi-automated computing procedure: (1) Using the Web Plot Digitizer approach, which has been shown to be valid and reliable [26], to import unstructured 2-D images. (2) Establishing the coordinate axes. (3) Points are chosen and data is retrieved for scatter data. (4) For curve data, comparable methods like X Step with Interpolation or Averaging Window automatically extract data points on the curve after masking [27].

Since the units and ranges of data that is acquired from various sources varies, z-normalization is used to standardize continuous input parameters:

$$Z_{i,j} = \frac{X_{i,j} - \mu_j}{\sigma_j} \tag{1}$$

where $\mu_j$ is the mean of the jth input variables, σj is the standard deviation of *j*th input variables, and $Z_{i,j}$ is the standard value of $X_{i,j}$, which is the jth input value of the *i*th data instance:

$$\sigma_j = \sqrt{\frac{1}{n-1} \sum_{k=1}^{n} (X_{k,j} - \mu_j)^2} \tag{2}$$

where the number of data instances is denoted by *n*. Each input variable's mean is about 0, its standard deviation is around 1, and the distribution of values is preserved when the z-normalization technique is used.

70% of the dataset is used for training, while 30% is used for testing. The dataset is divided into training and testing sets at random. Due to the small dataset size and the scarcity of test data for UHPC combinations in the literature, the splitting ratio was used.

**2.3 Predictive model selection**

2.3.1 Machine learning models

A total of 21 data-driven methods are compared in terms of the prediction accuracy of the compressive strength of UHPC. Nine conventional AI models are specifically examined: (1) Linear Regression, (2) Ridge Regression, (3) Ridge CV Regression, (4) L1-Regularized Linear Regression (Lasso), (5) Bayesian Linear Regression with Ridges, (6) Kernel Ridge, (7) Decision Tree, (8) ANN (Artificial Neural Network) (9) Support Vector Machine (SVM) Regression. Twelve hybrid or ensemble models are also implemented: (1) Adaptive Boosting (AdaBoost), (2) Bagging Regressor, (3) Light Gradient Boosting Machine (LightGBM), (4) Stacking Regressor, (5) Blending Ensemble, (6) Voting Regressor, (7) Least-Squares Boosting, (8) Random Forest, (9) Extra Tree Regressor, (10) Extreme Gradient Boosting (XGBoost), (11) Categorical Boosting (CatBoost), and (12) Gradient Boosting Decision Tree (GBDT). These models' performance in compressive strength prediction for UHPC is methodically assessed.

2.3.2 Performance indicators

To evaluate the efficacy of machine learning prediction models, six indicators are used: mean absolute error (MAE), mean square error (MSE), percentage mean absolute error (PMAE), root mean square error (RMSE), maximum absolute error (MaxAE), and coefficient of determination (R²). Equation (3) illustrates the anticipated values of various indicators. The ideas of these metrics are elaborated in Eq. (4)-(9), where $y_i$ denotes the genuine value and $\hat{y}_i$ signifies the anticipated value. To comprehensively evaluate the model's generalization performance and reduce the risk of overfitting, both training and testing outcomes are analyzed. A model is considered to possess strong generalization capability if it exhibits consistent performance across both sets, thereby signifying robustness and dependability.

$$Optimal_v = \min \begin{bmatrix} MAE \\ MAPE \\ MSE \\ RMSE \\ MaxAE \end{bmatrix}, \max(R^2) \tag{3}$$

$$MAE = \frac{1}{m} \sum_{i=1}^{m} |\hat{y}_i - y_i| \tag{4}$$

$$PMAE = \frac{1}{m} \sum_{i=1}^{m} \frac{|y_i - \hat{y}_i|}{|y_i|} \times 100 \tag{5}$$

$$MSE = \frac{1}{m} \sum_{i=1}^{m} (\hat{y}_i - y_i)^2 \tag{6}$$

$$RMSE = \left(\frac{1}{m} \sum_{i=1}^{m} (y_i - \hat{y}_i)^2\right)^{1/2} \tag{7}$$

$$MaxAE = \max(|y_i - \hat{y}_i|) \quad \forall\, i \in \{1, 2, , m\} \quad (8)$$

$$R^2 = 1 - \frac{\sum_{i=1}^{m}(y_i - \hat{y}_i)^2}{\sum_{i=1}^{m}(y_i - \bar{y})^2} \quad (9)$$

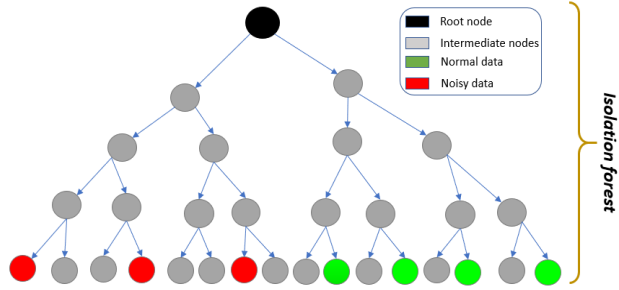

**Fig. 2**
Outlier detection using Isolation Forest. Black circle: root nodes of isolation trees. Dashed circles: investigated data points as intermediate nodes. Green circles: Normal data.

### 2.4 Outlier identification method

Datasets compiled from experimental studies and literature often include unusual data, also known as noise. Due to data processing, experimental, and collecting failures, noise data has outliers that exceed predicted results [28]. Outliers, which vary in distribution, attributes, or behavior, weaken regression-based prediction models [29]. Outliers are eliminated during data cleaning to improve model performance and decrease this issue. The 3-sigma rule, boxplot approach, and z-score test use statistical criteria to find outliers. Non-linear patterns and high-dimensional data challenge these approaches. This research uses the isolation forest technique to find outliers and clean data. To determine the degree of separation of data points, the isolation forest builds an ensemble of random binary trees, or isolation trees. Because outliers data points are easier to isolate from the bulk of the data, their average route lengths within these trees are often shorter. To identify outliers data, a threshold based on the contamination ratio is used; however, careful calibration is necessary to balance the removal of noise without excess shrinking the sample [30]. This technique has shown beneficial in managing large, high-dimensional datasets, and it has been used in earlier research to optimize concrete mix designs with the efficient detection of noise data [31]. This work intends to increase the prediction accuracy of machine learning models for UHPC property, flowability, and porosity assessment by integrating the isolation forest for data cleaning.

### 2.5 SHAP-based model interpretation

SHAP (SHapley Additive exPlanations) is a sophisticated method for understanding the results of machine learning prediction models. Calculating each input feature's prediction contribution shows how different qualities affect the model's decision-making. This strategy uses cooperative game theory's Shapley values to ensure each feature's fair contribution. Utilizing SHAP, scientists can decide which input parameters of the UHPC mix percentage have the most substantial influence on the performance prediction model of UHPC. Machine learning prediction model interpretability boosts integrity and user trust in its predictions.

### 2.6 Hyperparameter tuning

By assessing the model's accuracy on unknown data, K-fold cross-validation is applied to create a model with the required generalization performance. Fig. 3 shows K-fold cross validation flowchart. Randomization splits the training dataset into k folds of comparable sizes. After training using the remaining (k−1) folds, the model is tested on one of them. The trained model's loss function is evaluated. After each of the k folds is processed, the mean loss function is used to determine the model. The training dataset in this study is split into 10 folds (k = 10).

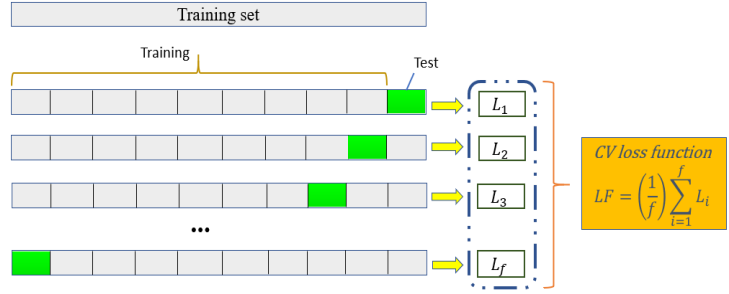

**Fig. 3**
K-fold cross-validation flowchart for model evaluation.

The objective of hyperparameter optimization by random search is to identify the ideal hyperparameter set.

$$\varphi^* = arg_{\varphi \in W} \min[RMSE(\varphi)] \qquad (10)$$

where $RMSE(\varphi)$ is the root mean square error associated with the hyperparameter set $\varphi$. The set $W$ comprises hyperparameter configurations randomly selected from the designated search space. Among them, $\varphi^*$ is the configuration that attains the minimal cross-validated RMSE, averaged over all folds.

## 3. Compiled database

Table 1 presents a detailed description of 17 unique mixture design parameters, 1,201 data points extracted from the literature on experimental testing of Ultra-High-Performance Concrete (UHPC) [32-50]. The 17 input features are used for developing a comprehensive machine learning model aimed at predicting key mechanical properties of UHPC, such as compressive strength, flexural strength and tensile strength. In addition, 14 input characteristics analyze flowability and porosity. This organized technique allows data-driven UHPC performance assessment. The compressive, flexural, and tensile strengths of UHPC 11, 14, and 15 are predicted using steel fiber content (SF), SF tensile strength (STL), and SF modulus of elasticity (MOE). Flowability and porosity predictions exclude slag powder, SF STL, and SF MOE. Table 1 shows the objective design parameters for compressive, flexural, tensile, flowability, and porosity: 18, 19, 20, 21, 22 respectively.

Table 1 Statistical summary of the variables of UHPC properties

| No. | Variable | Range | Unit | Mean | S.D. |
|---|---|---|---|---|---|
| 1 | Cement content | 369 - 1097 | Kg/m$^3$ | 797.21 | 170.77 |
| 2 | Coarse aggregate | 0 - 1931 | Kg/m$^3$ | 336.04 | 487.82 |
| 3 | Silica fume content | 0 - 279.2 | Kg/m$^3$ | 82.69 | 86.68 |
| 4 | Fly ash content | 0 – 301.76 | Kg/m$^3$ | 46.17 | 100.23 |
| 5 | Slag powder content | 0 – 468.9 | Kg/m$^3$ | 74.71 | 138.30 |
| 6 | Sand content | 0 - 1213 | Kg/m$^3$ | 808.42 | 320.61 |
| 7 | Superplasticizer | 0 – 88.2 | Kg/m$^3$ | 12.02 | 16.66 |
| 8 | Water content | 0 - 293 | Kg/m$^3$ | 172.20 | 46.04 |
| 9 | HPWR | 0 - 256 | Kg/m$^3$ | 20.59 | 34.00 |
| 10 | Water/binder ratio | 0 – 0.36 | % | 0.18 | 0.05 |
| 11 | Steel fiber content | 0 - 7 | % | 1.57 | 0.95 |
| 12 | Steel fiber diameter | 0 - 500 | μm | 159.75 | 126.91 |
| 13 | Steel fiber length | 0 - 30 | Mm | 11.80 | 7.27 |
| 14 | SF Tensile strength | 0 - 3842 | MPa | 1317.76 | 947.39 |
| 15 | SF Elastic modulus | 0 - 700 | GPa | 136.79 | 102.30 |
| 16 | Hydration Temperature | 0 - 26 | ℃ | 18.10 | 5.62 |
| 17 | Curing Age | 1 - 91 | Day | 20.54 | 22.68 |
| 18 | Compressive strength | 12.2 - 404 | MPa | 111.24 | 33.25 |

| 19 | Flexural strength | 0 – 36.2 | MPa | 7.30 | 9.39 |
| 20 | Tensile strength | 0 – 17.1 | MPa | 2.03 | 4.27 |
| 21 | Slump flow | 0 - 924 | Mm | 270.33 | 165.68 |
| 22 | Porosity | 0 - 18.6 | % | 2.69 | 4.77 |

## 4. Results and discussion

### 4.1 Analysis of multicollinearity within the dataset

Multicollinearity Reduction Method (MRM) has been used to decrease multicollinearity while ensuring feature interpretability, a correlation-based feature selection approach was used rather than dimensionality reduction methods like Principal Component Analysis (PCA). Pearson correlation coefficients were obtained for all input variables, and features with pairwise correlation ($|r| > 0.7$) were found, which does not affect model accuracy. Based on domain relevance and correlation strength with UHPC features, flowability, and porosity dataset, few variables were excluded from strongly linked pairs. This method reduced multicollinearity without impacting input factor physical significance. MRM model achieved accurate prediction with R² values of 0.953 and 0.923 for UHPC characteristics, flowability, and porosity. All input output feature Pearson correlation matrices are displayed in Fig. 4(a) & 4(b).

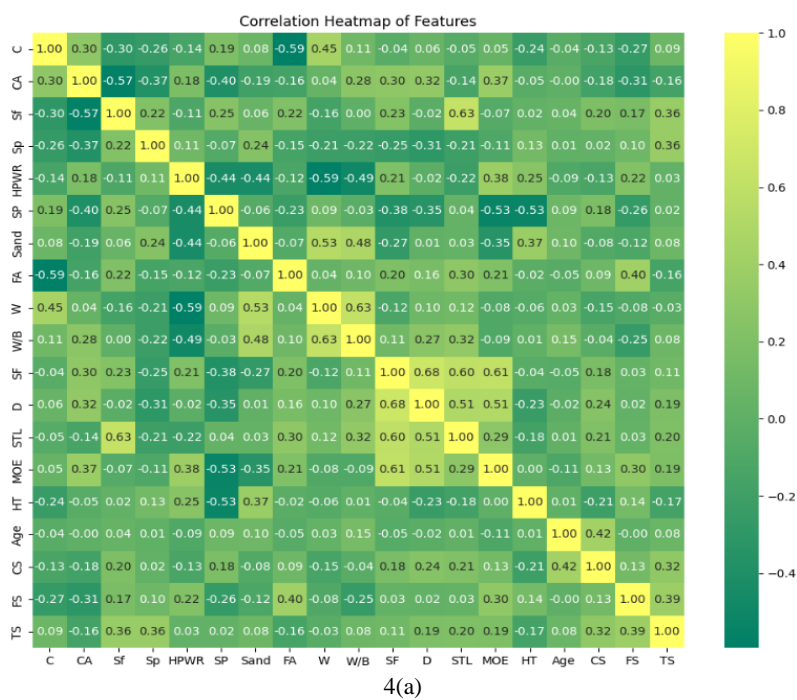

4(a)

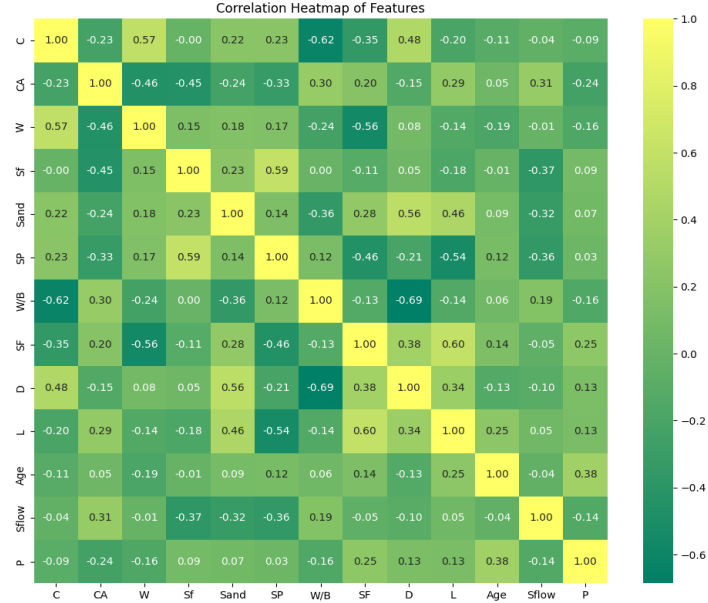

**Fig. 4**
Pearson correlation matrices of the features for predicting 4(a) compressive strength, flexural strength, tensile strength, 4(b) flowability, and porosity.

### 4.2 Pre-Selection of Machine Learning models

Table 2 lists the compressive strength prediction performance of 21 machine learning methods using default hyperparameters. The Bayesian Ridge models and Artificial Neural Network (ANN) have low accuracy, as shown by lower R² and higher RMSE values. Other models have good R² values (>0.80) for training but low R² for testing. We found five machine learning approaches with RMSE < 30 and R² > 0.18 for the testing dataset: Random Forest, Extra Tree Regressor, LightGBM, CatBoost, and XGBoost.

Table 2. Performance metrics of the machine learning algorithms

|   |   | Dataset | MAE | PMAE | MSE | RMSE | MaxAE | R² |
|---|---|---|---|---|---|---|---|---|
| 1 | Linear Regression | Training | 8.2 | 12.3 | 133.38 | 11.55 | 28.92 | 0.87 |
|   |   | Testing | 27.71 | 25.38 | 1879.56 | 43.35 | 323.4 | -0.71 |
| 2 | RidgeRegression | Training | 9.44 | 14.06 | 168.01 | 12.96 | 31.05 | 0.84 |
|   |   | Testing | 21.89 | 20.3 | 975.76 | 31.24 | 323.83 | 0.11 |
| 3 | RidgeCV | Training | 8.22 | 12.43 | 134.1 | 11.58 | 27.56 | 0.87 |
|   |   | Testing | 26.67 | 24.45 | 1662.77 | 40.78 | 323.47 | -0.51 |
| 4 | Lasso | Training | 9.41 | 13.97 | 168.22 | 12.97 | 31.14 | 0.84 |
|   |   | Testing | 21.86 | 20.26 | 973.53 | 31.2 | 323.77 | 0.11 |
| 5 | Support Vector Machine SVM | Training | 9.63 | 14.89 | 177.18 | 12.22 | 31.56 | 0.86 |
|   |   | Testing | 19.82 | 18 | 872.52 | 31.01 | 310.58 | 0.15 |
| 6 | BayesianRidge | Training | 24.32 | 39.41 | 986.79 | 31.41 | 89.22 | 0.05 |
|   |   | Testing | 26.78 | 25.33 | 1230.78 | 35.08 | 299.6 | -0.12 |
| 7 | KernelRidge | Training | 9.65 | 14.97 | 177.18 | 13.31 | 33.81 | 0.83 |
|   |   | Testing | 21.57 | 20 | 903.49 | 30.06 | 320.68 | 0.18 |
| 8 | DecisionTree | Training | 5.61 | 5.7 | 85.96 | 9.27 | 28.85 | 0.92 |
|   |   | Testing | 20.22 | 19.1 | 980.62 | 31.31 | 319.35 | 0.11 |
| 9 | AdaBoost | Training | 7.4 | 8.37 | 89.73 | 9.47 | 18.61 | 0.91 |

|  |  |  | Testing | 18.14 | 16.78 | 687.09 | 26.21 | 319.48 | 0.34 |
|---|---|---|---|---|---|---|---|---|---|
| 10 | BaggingRegressor |  | Training | 6.37 | 10.14 | 78.68 | 8.87 | 26.95 | 0.92 |
|  |  |  | Testing | 17.12 | 15.53 | 631.78 | 25.14 | 329.08 | 0.33 |
| 11 | LightGBM |  | Training | 15.75 | 25.18 | 430.53 | 20.75 | 62.12 | 0.59 |
|  |  |  | Testing | 18.81 | 17.92 | 684.9 | 26.17 | 306.05 | 0.38 |
| 12 | StackingRegressor |  | Training | 9.1 | 13.71 | 129.42 | 11.38 | 31.12 | 0.88 |
|  |  |  | Testing | 21.64 | 19.54 | 920.55 | 30.34 | 325.4 | 0.16 |
| 13 | BlendingEnsemble |  | Training | 5.79 | 8.36 | 56.18 | 7.5 | 32.08 | 0.95 |
|  |  |  | Testing | 12.11 | 11.69 | 372.36 | 19.3 | 342.39 | 0.35 |
| 14 | VotingRegressor |  | Training | 7.27 | 11.61 | 96.27 | 9.81 | 29.54 | 0.91 |
|  |  |  | Testing | 16.55 | 15.41 | 586.66 | 24.22 | 322.88 | 0.38 |
| 15 | ANN (Neural Network) |  | Training | 20.75 | 34.88 | 720.05 | 26.83 | 92.13 | 0.31 |
|  |  |  | Testing | 30.54 | 28.12 | 1605.31 | 40.07 | 295.54 | -0.46 |
| 16 | LeastSquaresBoosting |  | Training | 2.44 | 2.83 | 17.09 | 4.13 | 15.76 | 0.98 |
|  |  |  | Testing | 17.25 | 15.76 | 670.31 | 25.89 | 328.16 | 0.27 |
| 17 | **RandomForest** |  | Training | 6.33 | 10.03 | 76.49 | 8.75 | 26.49 | 0.93 |
|  |  |  | Testing | 17.18 | 15.62 | 629.59 | 25.09 | 328.97 | 0.40 |
| 18 | **ExtraTreeRegressor** |  | Training | 1.28 | 1.24 | 11.95 | 3.45 | 13.42 | 0.99 |
|  |  |  | Testing | 21.22 | 20.46 | 887.04 | 29.78 | 325.6 | 0.19 |
| 19 | **CatBoost** |  | Training | 2.13 | 2.48 | 13.82 | 3.72 | 14.43 | 0.99 |
|  |  |  | Testing | 18.62 | 16.93 | 733.34 | 27.08 | 326.04 | 0.33 |
| 20 | **GradientBoosting** |  | Training | 2.44 | 2.83 | 17.09 | 4.13 | 15.76 | 0.98 |
|  |  |  | Testing | 17.25 | 15.76 | 670.31 | 25.89 | 328.16 | 0.38 |
| 21 | **XGBoost** |  | Training | 1.28 | 1.24 | 11.95 | 3.46 | 13.43 | 0.99 |
|  |  |  | Testing | 17.32 | 16.13 | 666.85 | 25.82 | 333.6 | 0.39 |

## 4.3 Hyperparameter optimization

The five machine learning models' hyperparameters are tuned through random search. Table 3 lists ideal hyperparameters and ranges for compressive strength prediction. Random hyperparameter values within defined ranges are tested for each model, and the lowest cross-validated RMSE is chose. This approach guarantees efficient search space exploration and solid model performance. Table 4 presents compressive strength prediction performance metrics for the five techniques employing hyperparameter tuning. The XGBoost-based strategy has superior predictive performance of the five methods and is thus chosen for further analysis. Similar methods optimize UHPC's flexural, tensile, flowability, and porosity hyperparameters. XGBoost hyperparameters are best in Table 5. Optimization findings reveal that UHPC material properties greatly influence hyperparameter values. All five outputs' final XGBoost model results are in Table 6. These data show machine learning models require hyperparameter change. Therefore, machine learning models require numerous optimal hyperparameters for characteristics.

Table 3 The optimal hyperparameters for the five selected methods.

| Model | No. | Hyperparameter | Range | Optimal Value |
|---|---|---|---|---|
| RandomForest | 1 | n_estimators | 50 - 120 | 61 |
|  | 2 | max_depth | 10 - 30 | 14 |
|  | 3 | bootstrap | True - False | TRUE |
|  | 4 | min_samples_split | 10 - Feb | 2 |
|  | 5 | min_samples_leaf | 5 - Jan | 1 |
| XGBoost | 1 | learning_rate | 0.01 - 0.3 | 0.01 |
|  | 2 | max_depth | 2 - 20 | 18 |

|  |  | 3 | subsample | 0.5 - 1 | 0.94 |
|  |  | 4 | colsample_bytree | 0.5 - 1 | 0.94 |
|  |  | 5 | lambda_l1 | 0.05 - 1 | 0.55 |
|  |  | 6 | lambda_l2 | 0.05 - 1 | 0.67 |
|  |  | 7 | max_bin | 10 - 1000 | 20 |
|  |  | 8 | min_child_weight | 1 - 10 | 6 |
|  |  | 9 | gamma | 0 – 0.9 | 0.05 |
| CatBoost |  | 1 | learning_rate | 0.01 - 0.9 | 0.3 |
|  |  | 2 | depth | 10 - Apr | 4 |
|  |  | 3 | l2_leaf_reg | 5 - Jan | 1 |
|  |  | 4 | bagging_temperature | 0.1 - 2 | 0.5 |
|  |  | 5 | grow_policy | SymmetricTree - Depthwise | SymmetricTree |
|  |  | 6 | num_leaves | 31 - 63 | 31 |
|  |  | 7 | random_strength | 0.5 - 2 | 2 |
|  |  | 8 | subsample | 0.6 - 1.0 | 1 |
| ExtraTree |  | 1 | max_depth | 30 - Oct | 10 |
|  |  | 2 | min_samples_split | 10 - Feb | 2 |
|  |  | 3 | min_samples_leaf | 5 - Jan | 1 |
|  |  | 4 | n_estimators | 100 - 200 | 100 |
| GradientBoosting |  | 1 | learning_rate | 0.05 - 0.2 | 0.2 |
|  |  | 2 | n_estimators | 100 - 300 | 300 |
|  |  | 3 | max_depth | 10 - Mar | 3 |
|  |  | 4 | min_samples_split | 10 - Feb | 10 |
|  |  | 5 | min_samples_leaf | 5 - Jan | 1 |

Table 4 Performance metrics for the five selected methods (XGBoost has the best performance).

| Optimized model | Dataset | MAE | PMAE | MSE | RMSE | MaxAE | $R^2$ |
| --- | --- | --- | --- | --- | --- | --- | --- |
| Random Forest | Training | 6.75 | 7.61 | 66.21 | 8.12 | 19.88 | 0.93 |
|  | Testing | 5.85 | 5.69 | 75.75 | 8.7 | 38.78 | 0.94 |
| Gradient Boosting | Training | 5.48 | 6.68 | 52.65 | 7.26 | 14.52 | 0.94 |
|  | Testing | 5.92 | 5.8 | 82.49 | 9.08 | 52.09 | 0.93 |
| Extra Tree Regressor | Training | 9.49 | 11.48 | 136.36 | 11.68 | 29.17 | 0.85 |
|  | Testing | 7.78 | 7.19 | 181.84 | 13.48 | 99.2 | 0.85 |
| CatBoost | Training | 5.3 | 6.26 | 47.27 | 6.88 | 17.46 | 0.95 |
|  | Testing | 5.7 | 5.6 | 105.08 | 10.25 | 76.07 | 0.92 |
| **XGBoost** | Training | 1.83 | 2.09 | 5.23 | 2.29 | 6.12 | 0.96 |
|  | Testing | 5.65 | 5.61 | 74.27 | 8.62 | 43.52 | 0.94 |

Table 5 Final hyperparameter of XGBoost for five outputs

| No. | Hyperparameter | Range | Compressive strength | Flexural strength | Tensile strength | Flowability | Porosity |
| --- | --- | --- | --- | --- | --- | --- | --- |
| 1 | learning_rate | 0.01 - 0.3 | 0.01 | 0.15 | 0.06 | 0.08 | 0.14 |
| 2 | max_depth | 2 - 20 | 18 | 20 | 6 | 18 | 11 |
| 3 | subsample | 0.5 - 1 | 0.94 | 0.8 | 0.85 | 0.7 | 0.97 |
| 4 | colsample_bytree | 0.5 - 1 | 0.94 | 0.9 | 0.52 | 0.62 | 0.67 |

| 5 | lambda_l1 | 0.05 - 1 | 0.55 | 0.16 | 0.06 | 0.85 | 0.6 |
| 6 | lambda_l2 | 0.05 - 1 | 0.67 | 0.8 | 0.79 | 0.38 | 0.27 |
| 7 | max_bin | 10 - 2000 | 1000 | 1500 | 200 | 240 | 120 |
| 8 | min_child_weight | 1 - 10 | 6 | 2 | 5 | 4 | 5 |
| 9 | gamma | 0 - 0.9 | 0.05 | 0.05 | 0.5 | 0.7 | 0.9 |

Table 6 Performance of the final model with metrics of UHPC properties.

| Optimized model | Dataset | MAE | PMAE | MSE | RMSE | MaxAE | $R^2$ |
|---|---|---|---|---|---|---|---|
| Compressive strength | Training | 1.83 | 2.09 | 5.23 | 2.29 | 6.12 | 0.96 |
|  | Testing | 5.65 | 5.61 | 74.27 | 8.62 | 43.52 | 0.94 |
| Flexural strength | Training | 0.37 | 5.49 | 0.65 | 0.80 | 4.66 | 0.99 |
|  | Testing | 0.63 | 8.23 | 2.50 | 1.58 | 9.78 | 0.97 |
| Tensile strength | Training | 0.39 | 9.15 | 1.27 | 1.13 | 10.22 | 0.93 |
|  | Testing | 0.47 | 8.77 | 1.60 | 1.26 | 10.26 | 0.91 |
| Flowability | Training | 1.78 | 2.14 | 16.41 | 7.14 | 14.93 | 0.91 |
|  | Testing | 5.93 | 5.04 | 15.82 | 7.73 | 16.83 | 0.90 |
| Porosity | Training | 0.80 | 9.67 | 1.08 | 1.04 | 5.19 | 0.92 |
|  | Testing | 1.33 | 16.97 | 3.12 | 1.76 | 3.06 | 0.87 |

## 4.4 Outlier detection and SHAP analysis results

The unsupervised Isolation forest approach was used to discover outliers to strengthen prediction models. Compressive strength, flexural strength, tensile strength, flowability, and porosity were contaminated 10%, and outlier data was eliminated. When Model 2 was tested, this step improved prediction accuracy.

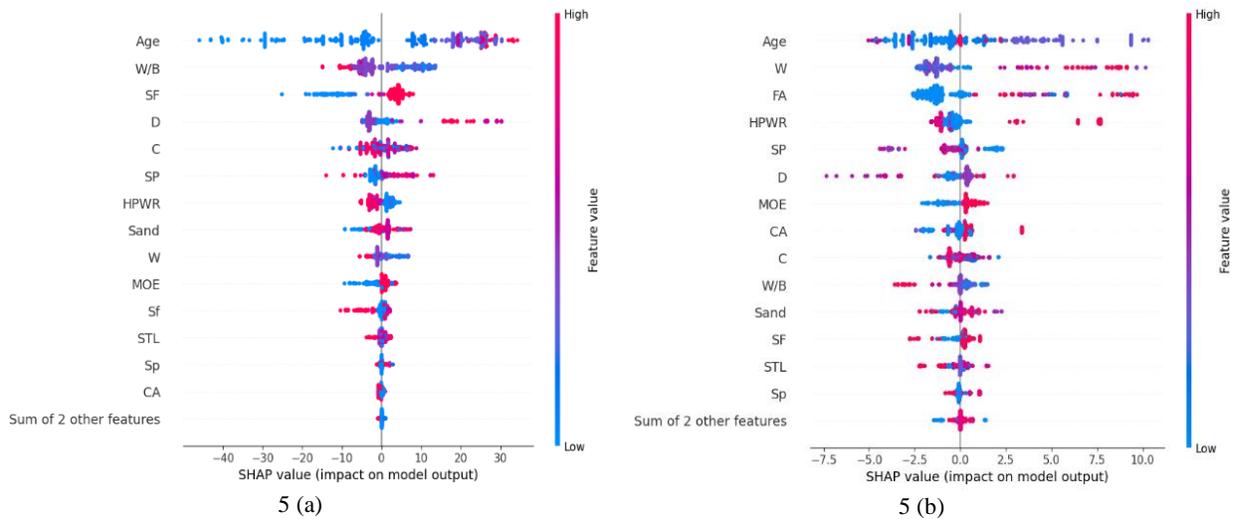

5 (a)  5 (b)

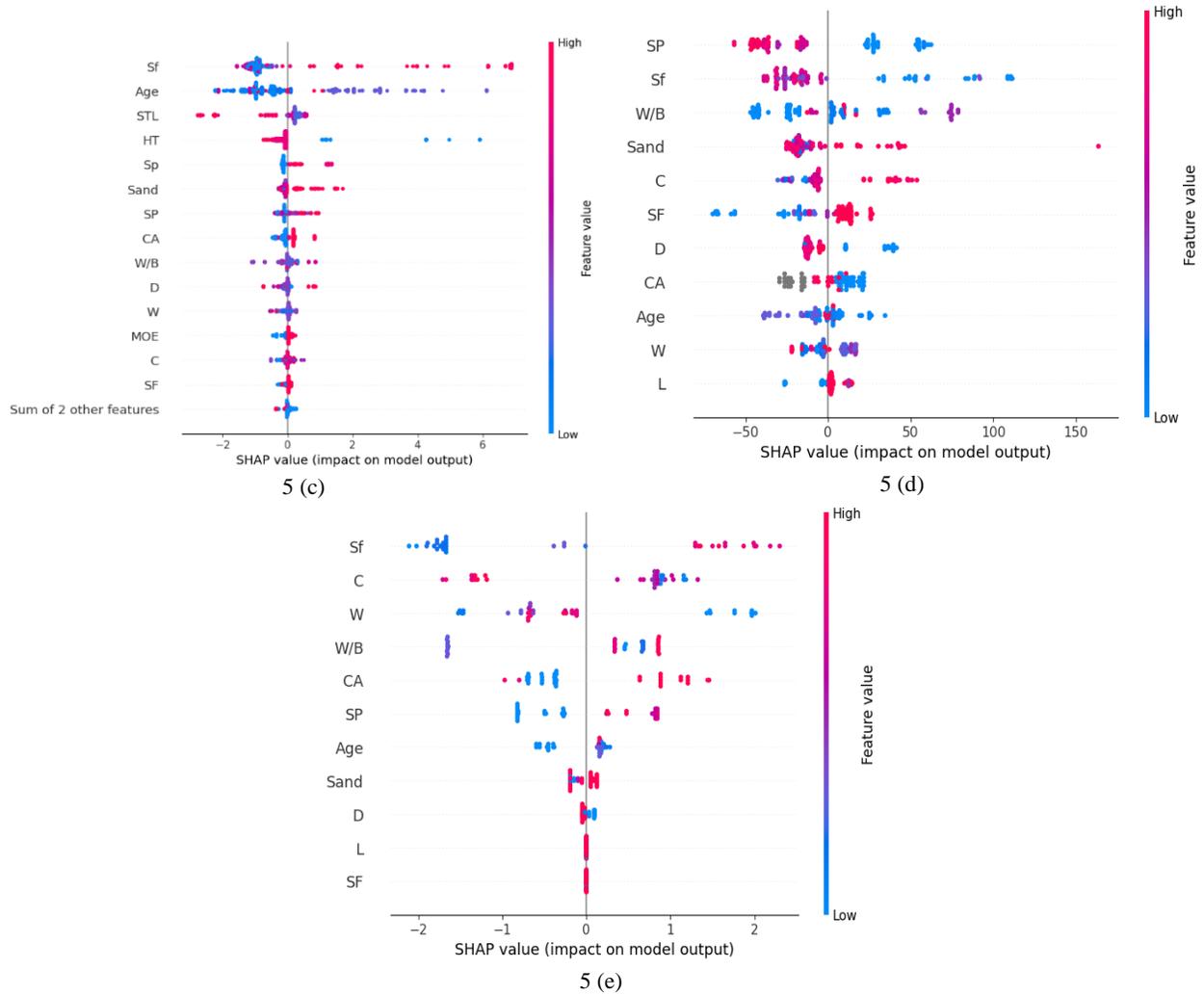

**Fig. 5**
SHAP interpretability evaluation of the UHPC regression prediction model (a). compressive strength (b). Flexural strength (c). Tensile strength, (d) Flowability, and (e) Porosity.

In Fig. 5(a), SHAP analysis indicates that curing age (Age), water-to-binder ratio (W/B), and steel fiber content (SF) effect UHPC compressive strength prediction. The curing age and steel fiber content improve compressive strength, whereas the water-to-binder ratio and water content diminish it. Hydration temperature, fly ash, and coarse aggregate have minimal impact. The key indicators of flexural strength (FS) of the model shows that curing age, water content (W), and fly ash (FA), a higher value improving performance. Additionally, cement (C) and super-plasticizers (SP) slightly increase FS at large amounts. High sand and HPWR impair flexural performance. SHAP tensile strength data shows that silica fume (Sf), curing age, and steel fiber tensile strength (STL) are most important. Superplasticizer (SP) and steel fiber may lower tensile strength, although curing age and silica fume boost it. Water and steel fiber diameter have minor but favorable effects. Superplasticizer (SP), water-to-binder ratio (W/B), and silica fume predominantly influence the flowability prediction model, with higher values boosting slump flow. Reduced coarse aggregate and free water limit flowability when cement and steel fiber percentage rises. Silica fume, cement, and water content greatly impact porosity. The study found that high water-to-binder ratio and superplasticizer concentration improve porosity via enhancing matrix density and packing. Higher sand and steel fiber content may reduce porosity owing to heterogeneous compaction or paste composition. SHAP interpretability matches UHPC material composition and physical behavior. Regression models are reliable and strong.

## 4.5 Impact of Mixture Design Features

Table 7 shows how mixture design factors affect model 2. High-impact SHAP interpretation analysis characteristics were evaluated and low-impact features were discarded. For compressive strength prediction, coarse aggregate (CA), fly ash (FA), and hydration temperature (HT) have low contribution; for flexural strength, silica fume (Sf), slag powder (Sp), and HT have lower performance; for tensile strength prediction, coarse aggregate (CA), fly ash (FA), and HPWR are not essential; A few characteristics are removed, reducing model prediction accuracy owing to multicollinearity.

Table 7 Selected importance features on mixture design for each output

| Features | Compressive strength (CS) | Flexural strength (FS) | Tensile strength (TS) | Flowability (Sflow) | Porosity (P) |
|---|---|---|---|---|---|
| Cement content (C) | Included | Included | Included | Included | Included |
| Coarse aggregate (CA) | Excluded | Included | Excluded | Included | Included |
| Silica fume content (Sf) | Included | Excluded | Included | Included | Included |
| Fly ash content (FA) | Excluded | Included | Excluded | Excluded | Excluded |
| Slag powder content (Sp) | Included | Excluded | Included | Excluded | Excluded |
| Sand content (Sand) | Included | Included | Included | Included | Included |
| Superplasticizer (SP) | Included | Included | Included | Included | Included |
| Water content (W) | Included | Included | Included | Included | Included |
| HPWR | Included | Included | Excluded | Excluded | Excluded |
| Water/binder ratio (w/b) | Included | Included | Included | Included | Included |
| Steel fiber content (SF) | Included | Included | Included | Included | Excluded |
| Steel fiber diameter (D) | Included | Included | Included | Included | Included |
| Steel fiber length (L) | Excluded | Excluded | Excluded | Excluded | Included |
| SF Tensile strength (STL) | Included | Included | Included | Excluded | Excluded |
| SF Elastic modulus (MOE) | Included | Included | Included | Excluded | Excluded |
| Hydration Temperature (HT) | Excluded | Excluded | Included | Excluded | Excluded |
| Curing Age | Included | Included | Included | Included | Included |

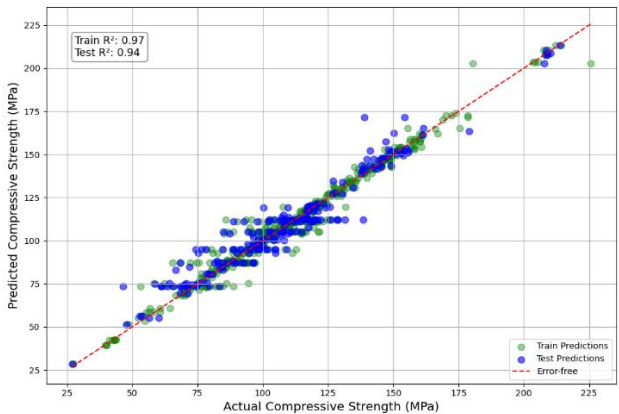

6 (a)

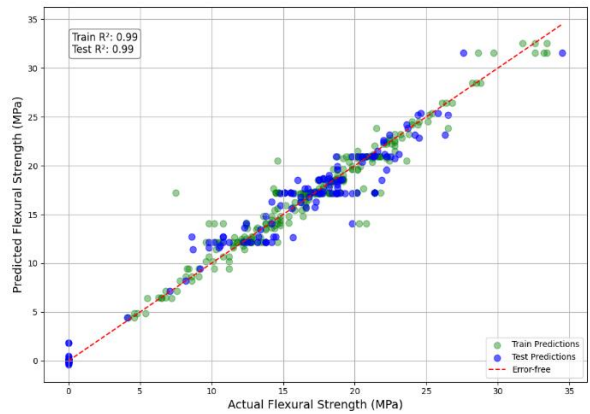

6 (b)

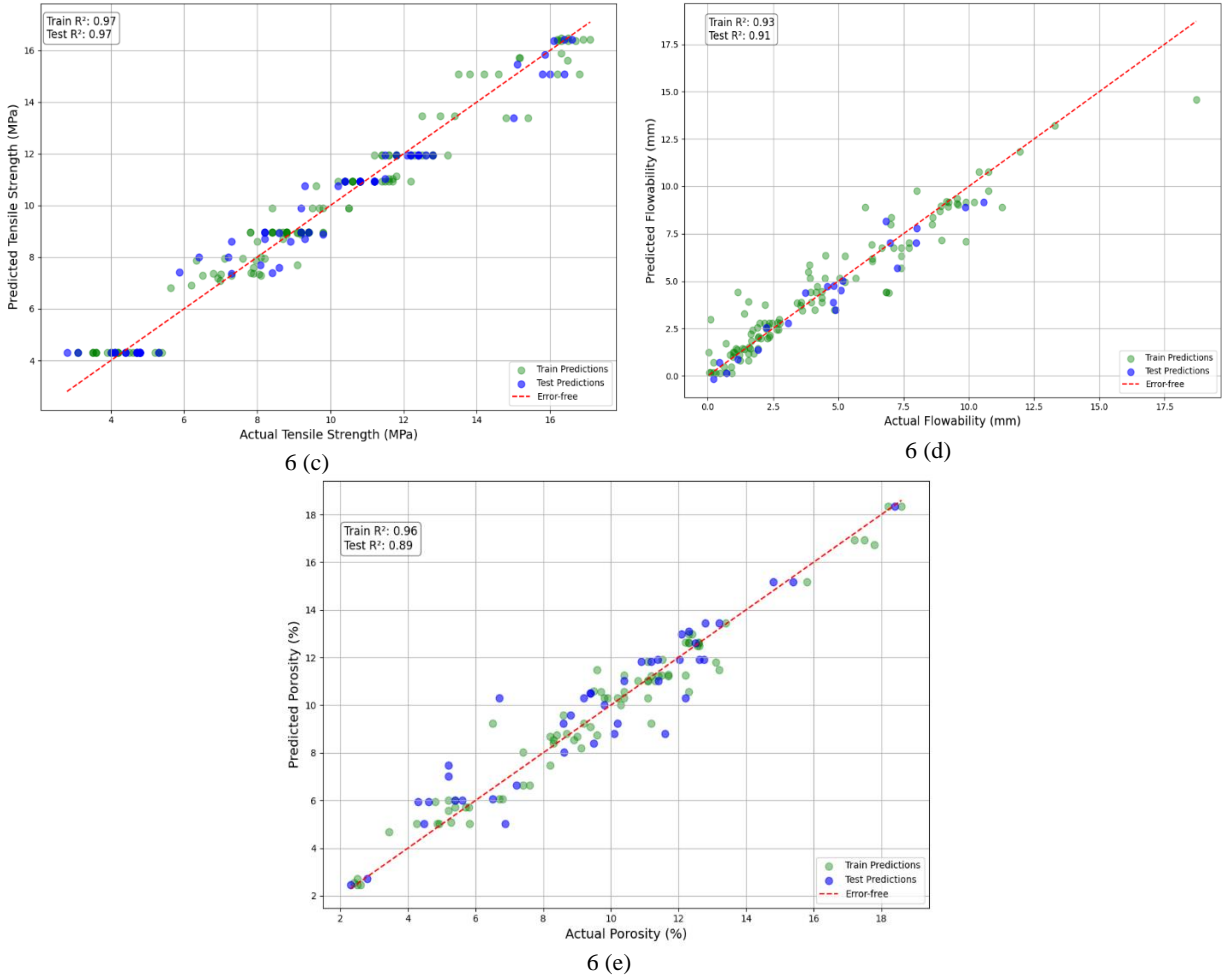

**Fig. 6**
The predicted results vs test outcomes for: (a) compressive strength, (b) flexural strength, (c) tensile strength, (d) flowability, and (e) porosity.

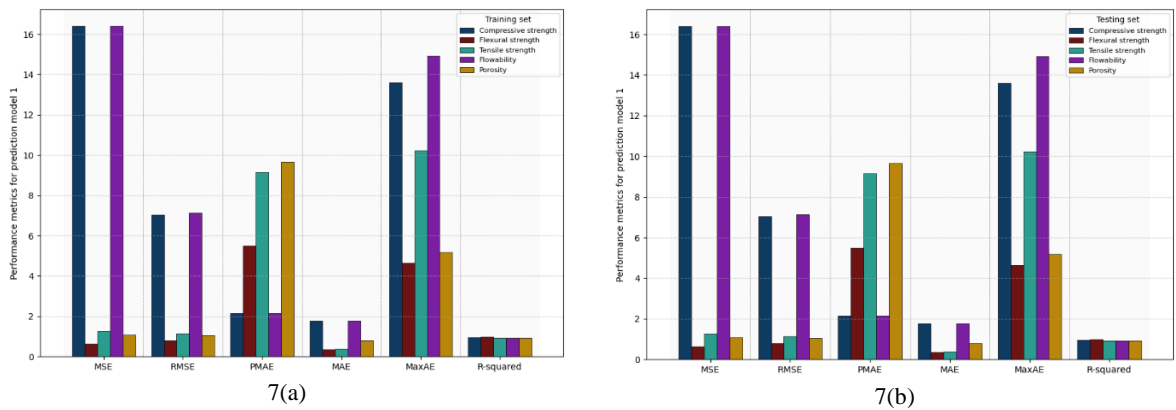

7(a)                7(b)

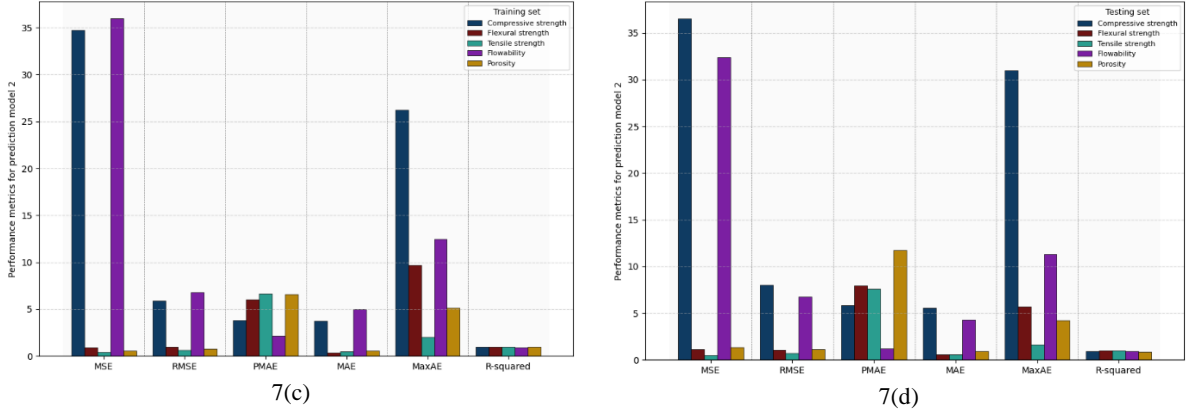

| | |
|---|---|
| 7(c) | 7(d) |

**Fig. 7**
The performance metrics for prediction model 1: 7 (a) training, 7 (b) testing, and model 2: (c) training, (d) testing

### 4.6 Prediction Results of UHPC Properties

Fig. 6 compares UHPC compressive, flexural, tensile, flowability, and porosity predictions to actual data. Overall, the five machine learning models had good prediction accuracy in training and testing. The minimal $R^2$ value for UHPC porosity testing is 0.87. Table 6 shows comprehensive results. Table 8 shows the performance metrics of the auto-tuned XGBoost, trained on a fresh dataset with no redundant data or low-impact features, showing excellent predicted accuracy.

Five machine learning models are developed and evaluated for each of the five UHPC target features to evaluate the influence of outlier identification and low-impact features selection on the predicted accuracy of auto-tuned XGBoost models:

(1) Model 1: auto-tuned XGBoost trained on the original dataset; (2) Model 2: refined by removing outlier data and extraneous features. Table 8 compares the performance metrics of the five machine learning models for each UHPC characteristic assessed in Fig. 7. Performance measurements show that Model 2 has the highest prediction accuracy than Model 1, proving that features selection and outlier identification improve prediction accuracy. The test dataset prediction accuracy has also increased, demonstrating better generalization.

Table 8 Performance metrics of the different models in predicting UHPC properties

| UHPC Properties | Prediction model | Dataset | MAE | PMAE | MSE | RMSE | MaxAE | $R^2$ |
|---|---|---|---|---|---|---|---|---|
| Compressive strength | Model 1 | Training | 1.78 | 2.14 | 16.41 | 7.05 | 13.59 | 0.96 |
| | | Testing | 5.93 | 5.04 | 59.82 | 11.73 | 28.83 | 0.94 |
| | Model 2 | Training | 3.75 | 3.84 | 34.75 | 5.89 | 26.26 | 0.97 |
| | | Testing | 5.56 | 5.84 | 36.52 | 8.01 | 30.99 | 0.94 |
| Flexural strength | Model 1 | Training | 0.37 | 5.49 | 0.65 | 0.80 | 4.66 | 0.99 |
| | | Testing | 0.63 | 8.23 | 2.50 | 1.58 | 9.78 | 0.97 |
| | Model 2 | Training | 0.38 | 6.05 | 0.93 | 0.97 | 9.72 | 0.99 |
| | | Testing | 0.55 | 7.92 | 1.16 | 1.08 | 5.73 | 0.99 |
| Tensile strength | Model 1 | Training | 0.39 | 9.15 | 1.27 | 1.13 | 10.22 | 0.93 |
| | | Testing | 0.47 | 8.77 | 1.59 | 1.26 | 10.26 | 0.91 |
| | Model 2 | Training | 0.53 | 6.66 | 0.43 | 0.66 | 2.00 | 0.97 |
| | | Testing | 0.55 | 7.61 | 0.48 | 0.69 | 1.60 | 0.97 |
| Flowability | Model 1 | Training | 1.78 | 2.14 | 16.41 | 7.14 | 14.93 | 0.91 |
| | | Testing | 5.93 | 5.04 | 15.82 | 7.73 | 16.83 | 0.90 |
| | Model 2 | Training | 4.97 | 2.19 | 35.98 | 6.78 | 12.46 | 0.93 |

|          |         | Testing  | 4.32 | 1.24  | 32.37 | 6.76 | 11.34 | 0.91 |
|----------|---------|----------|------|-------|-------|------|-------|------|
| Porosity | Model 1 | Training | 0.80 | 9.67  | 1.08  | 1.04 | 5.19  | 0.92 |
|          |         | Testing  | 1.33 | 16.97 | 3.12  | 1.76 | 3.06  | 0.87 |
|          | Model 2 | Training | 0.54 | 6.62  | 0.56  | 0.75 | 5.13  | 0.96 |
|          |         | Testing  | 0.90 | 11.76 | 1.37  | 1.17 | 4.23  | 0.89 |

## 4.7 Out-of-Set Validation

The study used literature-based experimental datasets for out-of-set validation of prediction models. Five non-training datasets were chosen. The compressive strength model was tested by comparing experimental data from ref-1 [51] to anticipated values. The largest absolute prediction error is 12.4%, as shown in Fig. 8(a).

To validate the flexural strength model, experimental data from ref-2 [52] was compared to projected values. As demonstrated in Fig. 8(b), predicts have a maximum absolute percentage error of 3.1%.

To validate the tensile strength model, experimental data from ref-3 [53] was compared to projected values. According to Fig. 8(c), the estimates' maximum absolute percentage error is -2.6%.

To assess the flowability model, the experimental data from ref-4 [54] was used for comparison with the predicted values. As shown in Fig. 8(d), the maximum absolute percentage error observed in the predictions is -8.3%.

To assess the porosity model, the experimental data from ref-5 [55] was compared with the predicted values. As shown in Fig. 8(e), the maximum absolute percentage error observed in the predictions is 8.6%.

These results indicate that most predicted outcomes are within an acceptable error range, with all maximum errors below 13%. This shows that the data-driven model has excellent predicted accuracy and generalization, enabling it to comprehend data beyond its training set.

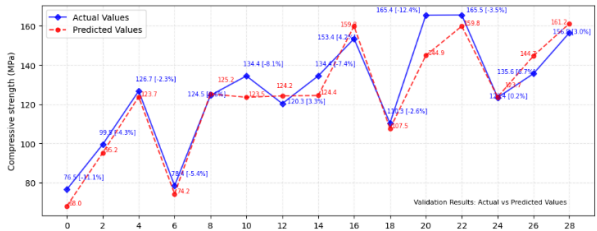
8(a)

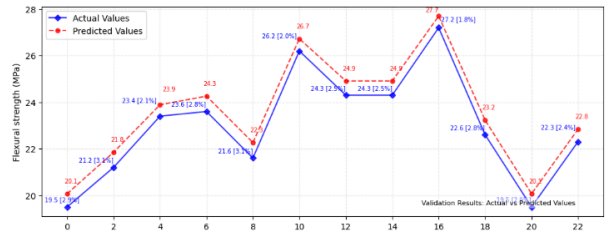
8(b)

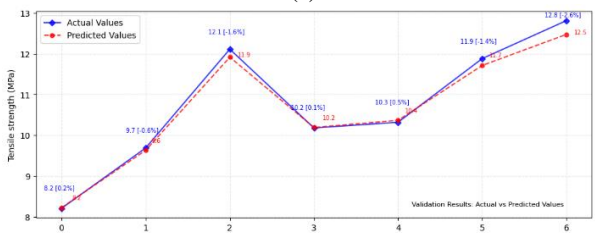
8(c)

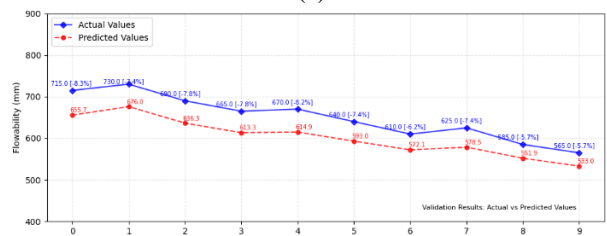
8(d)

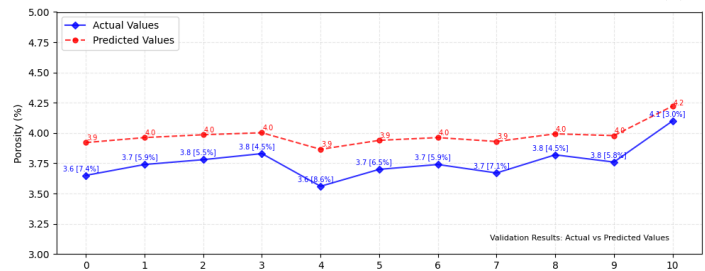

8(e)

**Fig. 8**
Out-of-set experimental data to evaluate the model's predicting efficacy for: (a) compressive strength, (b) flexural strength, (c) tensile strength, (d) flowability, and (e) porosity.

## 5. Conclusions and future scope

This research establishes that a new data-driven framework can reliably predict UHPC's compressive, flexural, tensile, flowability, and porosity with high accuracy and generality. To predict UHPC qualities, XGBoost was selected among 21 data-driven algorithms and used with automated hyperparameter tuning, k-fold cross-validation, outlier detection, and SHAP interpretation for feature significance selection. All of these evaluations provide these conclusions:

1. The data-driven framework accurately predicts compressive, flexural, tensile, flowability, and porosity of UHPC with high generalization capabilities. The prediction results from the test dataset showed RMSE values of 8.01, 1.08, 0.69, 6.76, and 1.17, with $R^2$ values of 0.94, 0.99, 0.97, 0.91, and 0.89. The framework helps design and optimize UHPC in real-world infrastructure projects because to its excellent predictive performance.
2. Integrity of the dataset used for training machine learning algorithms greatly impacts prediction accuracy. Model 2's feature selection and outlier identification improved machine learning model prediction. The prediction results using the test dataset showed that removing outlier data and selecting SHAP features for compressive strength, flowability, and porosity reduced RMSE values from 11.73 to 8.01, 7.73 to 6.76, and 1.76 to 1.17. Furthermore, compressive strength $R^2$ values maintained at 0.94, but flowability and porosity values rose from 0.90 to 0.91 and 0.87 to 0.89, respectively.
3. Model selection using random search and k-fold cross-validation may improve prediction accuracy for models like random forest, extra tree regressor, LightGBM, CatBoost, and XGBoost. Model 2 can be used to study how mixture design affects UHPC compressive, flexural, tensile, flowability, and porosity. Therefore, the models may speed up UHPC development by reducing testing.
4. Finally, A Web Application with a private Graphical User Interface (GUI) has been created for material designers in construction industry. It can be seen in the Fig.9  https://aa5b5488e330f9b839.gradio.live/

This research suggests the following areas for additional study:

   1. To improve predictability and interpretability, future research should use explainable AI methods like deep learning. The XGBoost model proved accurate. Improving model interpretability would enhance user trust and encourage material designers and construction engineers to employ it.

   2. Machine learning and LCA can assess UHPC compositions' economic and environmental impacts. Green infrastructure materials would be simpler to use.

   3. Future studies may study UHPC behavior using machine learning and multiscale modeling. UHPC prediction and scientific understanding may benefit from microstructural-macroscopic connections.

   4. Implementing this research's data-driven UHPC material manufacturing technique without AI is difficult. More research is needed to provide practical engineers a simple tool to develop UHPC for particular applications.

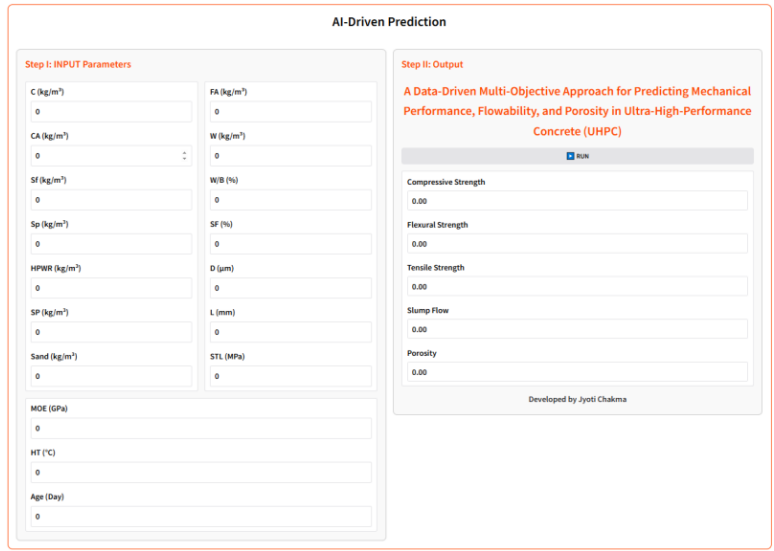

**Fig. 9**
Graphical user interface (GUI)

**CRediT authorship contribution statement**

**Chakma Jagaran:** Investigation, Data curation, Methodology, Data generation, Writing – original draft. **ZhiGuang Zhou:** Methodology, Writing – original draft, Conceptualization, Funding acquisition. **Chakma Jyoti:** Methodology, Writing – review & editing, Validation, Investigation, Data curation. **Cao YuSen:** Writing – review & editing, Investigation, Data curation.

**Declaration of competing interest**

The authors declare that they have no known competing financial interests or personal relationships that could have appeared to influence the work reported in this paper.

**Data availability**

Data will be made available on request.

**Acknowledgements**

The authors wish to gratefully acknowledge the support of this work by the National Key Research and Development Program of China under Grant No. 2024YFC3015100 and the National Natural Science Foundation of China under Grant No. 52378529.